\documentclass{article}

\usepackage{arxiv}

\usepackage[utf8]{inputenc} 
\usepackage[T1]{fontenc}    
\usepackage{hyperref}       
\usepackage{url}            
\usepackage{booktabs}       
\usepackage{amsfonts}       
\usepackage{nicefrac}       
\usepackage{microtype}      
\usepackage{lipsum}		
\usepackage{graphicx}
\usepackage{natbib}
\usepackage{doi}

\title{Viz: A QLoRA-based Copyright Marketplace for Legally Compliant Generative AI}

\author{ \href{https://orcid.org/0000-0001-5431-6367}{\includegraphics[scale=0.06]{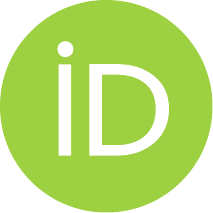}\hspace{1mm}Dipankar Sarkar} \\
  Terraprompt AI \\
  \texttt{me@dipankar.name} \\
}

\hypersetup{
pdftitle={Viz: A QLoRA-based Copyright Marketplace for Legally Compliant Generative AI},
pdfsubject={cs.LG},
pdfauthor={Dipankar Sarkar},
pdfkeywords={Copyright,LLM,QLoRa,Economics,Marketplace},
}

\begin{document}
\maketitle

\begin{abstract}
This paper aims to introduce and analyze the Viz system in a comprehensive way, a novel system architecture that integrates Quantized Low-Rank Adapters (QLoRA) to fine-tune large language models (LLM) within a legally compliant and resource efficient marketplace. Viz represents a significant contribution to the field of artificial intelligence, particularly in addressing the challenges of computational efficiency, legal compliance, and economic sustainability in the utilization and monetization of LLMs. The paper delineates the scholarly discourse and developments that have informed the creation of Viz, focusing primarily on the advancements in LLM models, copyright issues in AI training \cite{NYT_OpenAI_2023}, and the evolution of model fine-tuning techniques, particularly low-rank adapters and quantized low-rank adapters, to create a sustainable and economically compliant framework for LLM utilization.The economic model it proposes benefits content creators, AI developers, and end-users, delineating a harmonious integration of technology, economy, and law, offering a comprehensive solution to the complex challenges of today's AI landscape.
\end{abstract}

\keywords{Copyright \and LLM \and QLoRa \and Economics \and Marketplace}

\section{Introduction}

The realm of artificial intelligence (AI), particularly in the field of large language models (LLMs), has seen a substantial evolution in recent years, driven by innovations in model architectures, training methodologies, and application scopes. However, this rapid advancement brings to the fore significant challenges, notably in the domains of computational efficiency, economic viability, and legal and ethical concerns, particularly relating to copyright issues \citep{NYT_OpenAI_2023} in the pre-training of these models. This paper introduces Viz, a novel system architecture designed to address these challenges by leveraging recent advances in model fine-tuning techniques, specifically QLoRA (Quantized Low Rank Adapters), to create a sustainable and legally compliant framework for LLM utilization.

Language models, such as the GPT (Generative Pre-trained Transformer) series developed by OpenAI, have demonstrated remarkable capabilities in natural language understanding and generation, underpinning various applications from chatbots to content creation \citep{Radford2019LanguageMA}\citep{brown2020language}. Despite their impressive performance, these models often require extensive computational resources for training and fine-tuning, which poses significant economic and environmental concerns \citep{strubell-etal-2019-energy}. Furthermore, the pre-training of such models typically involves large datasets sourced from the Internet, raising critical copyright issues as witnessed in recent legal challenges \citep{GaonAviv2021Tfoc}.

The introduction of LoRA (Low-Rank Adaptation) presented a paradigm shift in the fine-tuning of large models, offering a more resource-efficient approach \citep{hu2021lora}. Building upon this, QLoRA emerges as a groundbreaking technique, allowing for the fine-tuning of models with tens of billions of parameters on comparatively modest hardware while maintaining performance levels. The QLoRA methodology, as elucidated in its seminal paper, introduces several innovative techniques such as 4-bit NormalFloat quantization and Double Quantization, which significantly reduce memory usage without sacrificing task performance \citep{dettmers2023qlora}. The Guanaco family of models, fine-tuned using QLoRA, demonstrates exemplary performance on the Vicuna benchmark, achieving near parity with state-of-the-art models like ChatGPT with markedly reduced resource requirements.

Viz leverages these advances to create a collaborative ecosystem involving content providers, AI developers, and end-users. At its core, Viz proposes a model where LLMs are initially pretrained on non-copyrighted datasets. Content providers can then utilize QLoRA to create fine-tuned models specific to their content, which are subsequently offered in a marketplace. This marketplace functions akin to digital content platforms like Spotify, allowing users to access and combine multiple fine-tunes for their specific needs, with usage tracked and monetized accordingly.

This paper aims to detail the architecture of Viz, its implementation using QLoRA, and the economic model it proposes, along with a thorough discussion of the legal, ethical, and practical implications of such a system. Through Viz, we envision an economically viable, legally compliant, and resource-efficient paradigm for the utilization and monetization of large-language models, heralding a new era in the field of artificial intelligence. Here are the contribution of the papers.

\begin{itemize}
    \item Innovatively integrates Quantized Low-Rank Adapters (QLoRA) within a marketplace framework, revolutionizing the accessibility and efficiency of large language models (LLMs).
    \item Provides a comprehensive design of the Viz system that addresses computational efficiency, copyright compliance, and economic sustainability in AI.
    \item Proposes a sustainable economic model for AI technology, ensuring legal and ethical compliance, crucial for responsible AI deployment.
    \item Contributes to the discussion on legal and ethical considerations in AI, particularly in copyright compliance and data privacy.
\end{itemize}

\section{Literature Review}

The foundation of the Viz system is based on the growing field of large-language models (LLMs) and their methods of finetuning, as well as the legal and ethical dilemmas associated with them. This review of existing literature aims to outline the academic discussions and progress that have influenced the development of Viz. The primary focus is on the advancements in LLMs, copyright concerns in AI training, and the evolution of fine-tuning techniques, specifically Low-Rank Adapters (LoRA) and Quantized Low-Rank Adapters (QLoRA).

\paragraph{Genesis and Evolution of LLMs} Beginning with the introduction of the Transformer architecture \citep{vaswani2023attention}, the field has seen a rapid expansion, most notably with models like BERT \citep{devlin2019bert} and GPT series \citep{Radford2019LanguageMA}\citep{brown2020language}. These models have revolutionized natural language processing (NLP), offering unprecedented capabilities in language understanding and generation.

The versatility of LLMs in various applications ranging from text generation to conversational AI has been extensively documented \citep{wolf2020huggingfaces}. Their impact extends beyond technical domains, influencing fields such as law, education, and creative industries.

\paragraph{Data Sourcing and Copyright Challenges} The reliance of LLMs on vast datasets, often scraped from the web, raises significant copyright concerns \citep{GaonAviv2021Tfoc}. The legal discourse around data usage for AI training is a burgeoning field, with scholars debating the balance between innovation and copyright protection \citep{10.1093/oso/9780190246693.001.0001}.

The necessity for adhering to data usage practices in line with regulations is underscored by recent legal actions and examination in the field of AI. The Oracle v. Google (2021) lawsuit sheds light on the intricate legal considerations associated with utilizing copyrighted content for software development \citep{enwiki:1187395454}.

\paragraph{Advent of LoRA} The introduction of LoRA by \citep{hu2021lora} marked a significant shift in fine-tuning practices. By applying low-rank matrices to modify pre-trained weights, LoRA offered a more efficient alternative to full model retraining, allowing for resource-efficient customization of LLMs.

QLoRA, an extension of LoRA, is a notable improvement in the efficiency of fine-tuning. The foundational paper describes its techniques, including 4-bit NormalFloat quantization and Double Quantization, which allow for the fine-tuning of larger models (such as the 65B parameter models) on constrained hardware while maintaining performance \citep{dettmers2023qlora}.

\paragraph{Economic Models in AI and Content Monetization}The monetization of AI technologies, especially in the realm of LLMs, presents unique challenges and opportunities. The work of \citep{10.7208/chicago/9780226613475.001.0001} on the economics of artificial intelligence provides a comprehensive overview of these dynamics.

The economic model proposed by Viz parallels existing digital content platforms. Studies on platforms like Spotify \citep{NBERw24713} offer insights into user-based monetization models, which are analogous to Viz's approach to monetizing fine-tuned LLMs.

This review highlights the dynamic nature of LLMs, which are characterized by rapid innovation and intricate challenges. The Viz system, positioned at the convergence of these advancements, seeks to leverage the capabilities of LLMs while tackling the computational, economic, and legal obstacles that are commonly encountered in this domain.

\section{Viz System Architecture}

The design of the Viz system is highly complex in order to promote a collaborative interaction among content providers, generative AI developers, and end-users. This is achieved by utilizing the latest advancements in fine-tuning large language models (LLMs) through Quantized Low Rank Adapters (QLoRA). 

\begin{figure}[ht]
    \centering
    \includegraphics[width=0.8\linewidth]{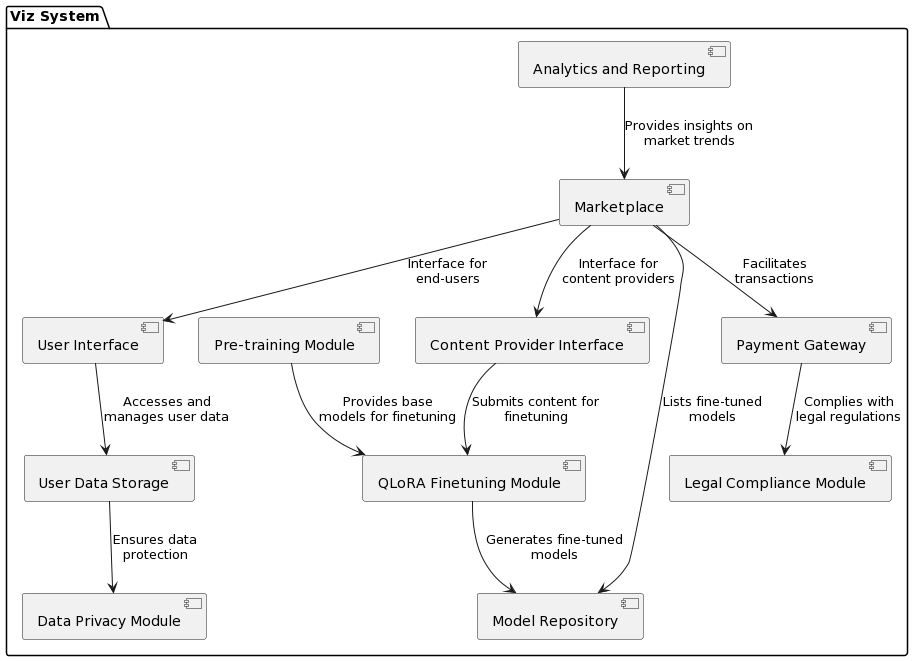}
    \caption{Viz system components}
    \label{fig:components}
\end{figure}

In this section, we will provide an overview of the key elements \ref{fig:components}, technical requirements, and general process of the Viz system, elucidating how it addresses the computational, financial, and legal challenges in the existing AI landscape.

\subsection{Fundamentals}

Viz is conceptualized as a platform that integrates a marketplace for AI models fine-tuned through QLoRA, providing a legally compliant and economically viable avenue for content creators and users to interact with LLMs.

The system aims to reduce computational overhead, ensure copyright compliance in training datasets, and create a sustainable economic model for all stakeholders.

\paragraph{Pre-training on Non-copyrighted Corpus} The initial stage involves training LLMs on a vast corpus of data that is free from copyright restrictions. This approach aligns with the legal frameworks discussed by \cite{GaonAviv2021Tfoc}, further addressing the challenges highlighted in Oracle v. Google (2021) \cite{enwiki:1187395454} and NYT case (2023) \cite{NYT_OpenAI_2023}.
\paragraph{QLoRA for Content-specific Fine-tuning} Building on the efficiency of QLoRA, as expounded by \cite{hu2021lora}, content providers can fine-tune models specific to their content. This process utilizes the innovations in QLoRA, such as 4-bit NormalFloat quantization, to manage large model parameters effectively.
\paragraph{Marketplace for Finetune Modules} A digital platform akin to \cite{NBERw24713} study on Spotify, where fine-tuned models are available for purchase or rent. This marketplace facilitates the monetization of AI models, aligning with the economic models proposed by \cite{10.7208/chicago/9780226613475.001.0001}.
\paragraph{User Interface for Module Application and Tracking} An intuitive interface allows end users to apply multiple fine-tunings to their LLM usage, with an integrated tracking system for usage monitoring and billing, reminiscent of digital content platforms.

\begin{figure}[ht]
    \centering
    \includegraphics[width=0.8\linewidth]{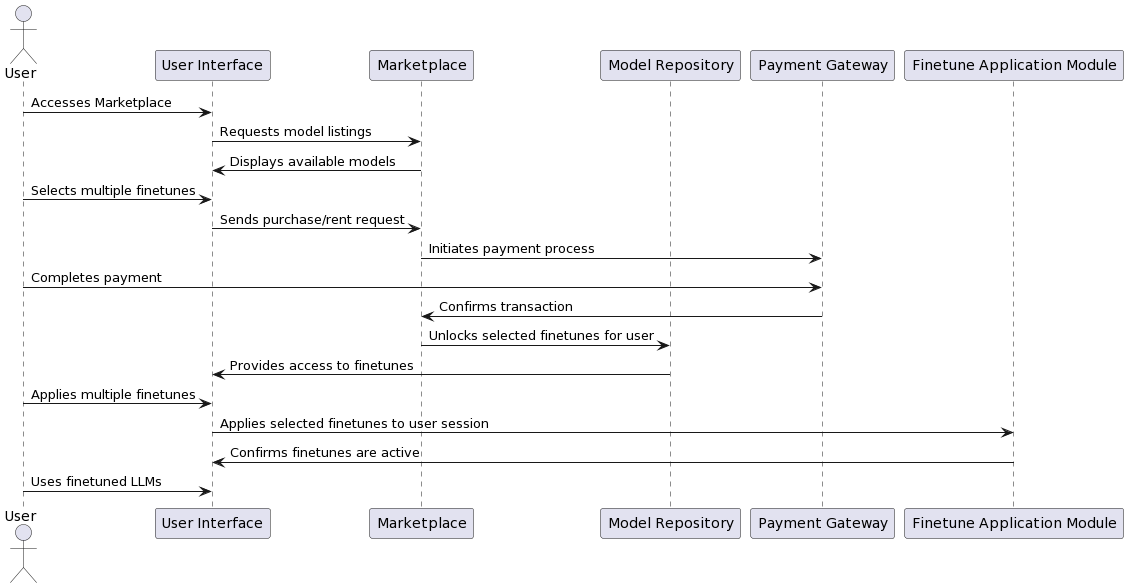}
    \caption{Consumer workflow}
    \label{fig:consumer}
\end{figure}

\begin{figure}[ht]
    \centering
    \includegraphics[width=1.0\linewidth]{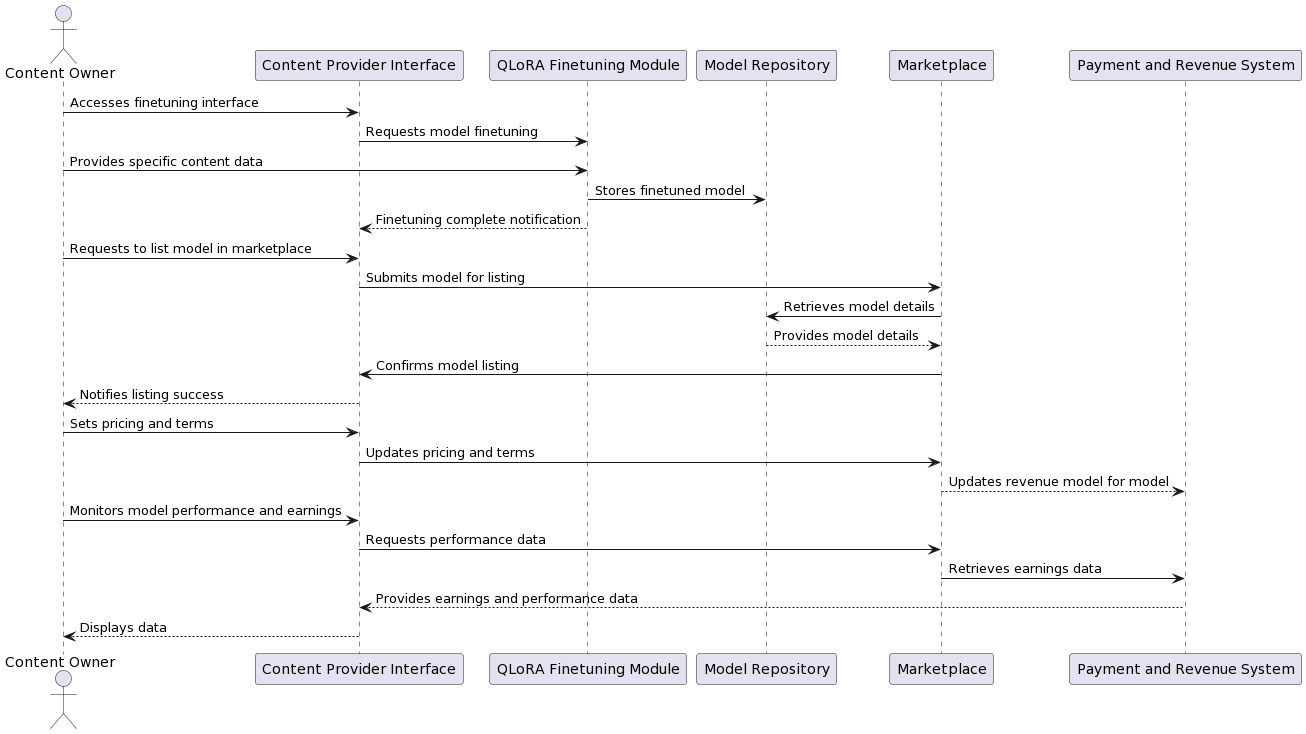}
    \caption{Content owner workflow}
    \label{fig:content-owner}
\end{figure}

Here \ref{fig:consumer} step-by-step description of the training of base LLMs and their subsequent fine-tuning using QLoRA.  We have \ref{fig:content-owner} explanation of how content providers upload and price their fine-tuned models and how users can access and utilize these models. It also details the mechanism for tracking the usage of fine-tunes and the subsequent billing process.

The design of the Viz system offers a holistic and inventive approach to the existing obstacles in the artificial intelligence field, specifically in the application of LLMs. Through the incorporation of advanced methods such as QLoRA for effective model refinement and the establishment of a marketplace that advantages all parties involved, Viz establishes a standard for upcoming advancements in the AI sector and the monetization of content.
\subsection{QLoRA importance in Viz}

The incorporation of Quantized Low Rank Adapters (QLoRA) into the Viz system architecture represents a notable progress in the efficient and successful fine-tuning of large language models (LLMs). This section thoroughly describes the application of QLoRA in Viz, clarifying the technical complexities and operational benefits that come with this integration, particularly in terms of resource utilization and performance improvement.

\paragraph{QLoRA's Core Principles} At the heart of QLoRA, as elaborated in its seminal paper, lies the concept of using low-rank matrices in conjunction with quantization techniques to fine-tune LLMs. This method significantly reduces the computational overhead traditionally associated with fine-tuning such models.

QLoRA introduces several key innovations, including 4-bit NormalFloat (NF4) quantization and Double Quantization, which collectively contribute to its memory efficiency. These techniques enable the fine-tuning of models with exceptionally large parameters (such as 65B) on limited hardware resources, aligning with the findings of \cite{hu2021lora}.

\paragraph{Adapting QLoRA for Viz} The implementation of QLoRA within Viz should be carefully tailored to align with the system's objective of providing a scalable and legally compliant platform for LLM utilization. This adaptation involves optimizing QLoRA's existing capabilities to suit the diverse requirements of content providers and end-users in the Viz marketplace.

In Viz, QLoRA is employed to create a modular fine-tuning framework. This approach allows content providers to fine-tune base LLMs with their specific datasets, leveraging QLoRA's efficiency to handle diverse content types and requirements.

\paragraph{Efficient Resource Utilization} By integrating QLoRA, Viz significantly lowers the barriers to entry for content providers, enabling them to fine-tune LLMs on comparatively modest hardware setups, thereby democratizing access to advanced AI capabilities.

\paragraph{Enhanced Model Performance} The QLoRA fine-tuned models in Viz, benchmarked on standards such as the Vicuna benchmark, demonstrate that this integration does not compromise on performance, ensuring that users have access to high-quality AI models.

The implementation of QLoRA within the Viz system represents a paradigm shift in the fine-tuning of LLMs, offering a solution that is both resource-efficient and performance-oriented. This integration not only enhances the capabilities of Viz as a platform for AI model monetization but also contributes significantly to the broader field of AI and machine learning by demonstrating a practical and scalable approach to model customization.

\section{Marketplace Design and Economics}

The Viz system integrates an innovative marketplace for distributing and earning money from finely-tuned large language models (LLMs). This marketplace utilizes the advancements made through Quantized Low Rank Adapters (QLoRA). In this section, we explore the architectural design of the marketplace, its economic foundations, and the proposed revenue models. We draw comparisons to existing digital platforms and economic theories in the field of digital goods and AI.

\subsection{Structural Design}
The digital platform in Viz is organized as a marketplace, where content providers can upload and offer their QLoRA fine-tuned LLMs for sale or lease. This structure is inspired by the model seen in digital content platforms such as Spotify, as discussed by \cite{NBERw24713}.

Models in the marketplace are categorized based on various criteria, such as application domain, language, and performance metrics, facilitating easy access and selection by end-users.

\subsection{Economic Model for Content Providers and Users}

The marketplace employs a dual monetization strategy. Content providers can opt to either sell their models outright or offer them on a subscription basis. This approach aligns with contemporary digital economic models as discussed by \cite{10.7208/chicago/9780226613475.001.0001}.

Pricing in the marketplace is dynamic, allowing content providers to set prices based on factors like model performance, uniqueness, and demand. The Viz system also incorporates algorithms to suggest optimal pricing strategies based on market trends.

\subsection{Revenue Sharing Models}
The income generated from the sales and subscriptions of models is divided between the content providers and the Viz platform, following a pre-established agreement for revenue sharing. This approach mirrors the revenue-sharing practices observed in digital marketplaces, as examined in research conducted by \cite{10.1162/154247603322493212}.

Viz incentivizes content providers through a reward system that recognizes high-performing and popular models, encouraging continuous improvement and innovation in model development.

\subsection{Comparison with Existing Models}
The economic structure of the Viz marketplace is compared with existing digital platforms, such as Spotify and Netflix, highlighting similarities and differences in terms of user engagement, pricing, and revenue models.

The marketplace's design is analyzed through the lens of economic theories pertinent to digital goods, such as those proposed by \cite{varian1995pricing}, focusing on aspects like pricing, copyright, and distribution in digital economies.

The marketplace within the Viz system represents a novel and economically viable avenue for the distribution and monetization of AI models. By incorporating advanced AI technologies like QLoRA and integrating sound economic principles, the marketplace not only fosters innovation in AI model development but also presents a sustainable economic model for content providers and users alike.

\section{Legal and Ethical Considerations}

Ensuring a deep comprehension and strict adherence to legal and ethical standards is of utmost importance in the creation and functioning of the Viz system, particularly due to the intricate nature of large language models (LLMs) and AI technology. This section highlights the various legal and ethical factors that are essential to the Viz system, with specific emphasis on copyright adherence, data privacy, AI ethics, and user safeguarding.

The primary objective of the Viz system is to guarantee adherence to global copyright regulations, especially when it comes to utilizing data for training LLMs. In light of recent legal disputes such as Oracle v. Google (2021) \citep{enwiki:1187395454}, which highlight the intricate legal aspects of incorporating copyrighted material into software, it is essential to ensure compliance. To ensure this, content providers in the Viz marketplace must verify that the data used for fine-tuning models through QLoRA does not violate any copyright laws, following the guidelines presented by \cite{GaonAviv2021Tfoc}.

The Viz system incorporates strong data security measures to protect user data, in accordance with international data privacy regulations like the General Data Protection Regulation (GDPR). The system's protocols for data privacy are designed to prevent unauthorized access and misuse of user data. Viz will maintain transparency regarding data usage, ensuring that users are informed about how their data is utilized within the system, as recommended in privacy-focused literature \citep{cavoukian2009privacy}.

The Viz system subscribes to the principles of ethical AI, including fairness, accountability, and transparency. These principles are critical in mitigating biases and ensuring equitable AI outcomes, as discussed by \citep{dignum2019responsible}. Viz implements a governance framework for AI models in the marketplace, overseeing their development and deployment to ensure they adhere to ethical standards and do not propagate harmful content or biases.

Viz is committed to the principles of fair use, which means that it values the intellectual property rights of others when using models and their outputs. This is crucial for maintaining a legally compliant and ethically responsible AI ecosystem. Viz has implemented measures within its system to prevent the inappropriate use of AI technology, such as creating misleading or harmful content. These measures align with the ethical guidelines put forth by \cite{jobin2019global}.

The economic models employed in the Viz marketplace are scrutinized for legal implications, ensuring that the monetization of AI models aligns with copyright and trade laws. The marketplace adheres to ethical monetization practices, promoting fairness and avoiding exploitation, as outlined in the economic and ethical discourse by \cite{lanier2014owns}.

The legal and ethical framework of the Viz system is a cornerstone of its design and operation, ensuring that it not only advances technological innovation but also upholds the highest standards of legality, ethics, and user protection. This framework is instrumental in building trust among users, content providers, and stakeholders in the AI community.

\section{Discussion}

In this section, we conduct a thorough examination of the Viz system, with a specific emphasis on its influence on the AI and content industry. We also explore the potential obstacles it may face and discuss its future advancements, such as the possibility of decentralization. The purpose of this discussion is to place the Viz system within the larger context of technological advancements, economic models, and evolving legal frameworks.

The Viz system, with its unique utilization of QLoRA and the marketplace model, is a notable advancement in the AI and content sector. However, for its future development, which may involve decentralization, it is crucial to navigate the technical, legal, and ethical aspects with caution. The system's progression will undoubtedly contribute to the ongoing discussions in the fields of AI, digital economy, and technology governance.

\subsection{Impact on AI and Content Industry}

By incorporating QLoRA, Viz effectively reduces the obstacles to developing and customizing AI models, making it more accessible to a wider group of individuals. This aligns with the notion put forth by \cite{benker2006wealth} in their research on the networked information economy, emphasizing the democratization of access.

The Viz marketplace model introduces a new economic paradigm in the AI industry, aligning with the shifts toward digital economies and platform-based models (\citep{parker2016platform}. It fosters a competitive environment where innovation is incentivized, potentially leading to rapid advancements in AI applications.

\subsection{Decentralization}

Implementing decentralized structures for data management could enhance security and user control over data, resonating with the principles of Web3.0.

A decentralized marketplace could offer greater transparency in transactions, fairer revenue distribution, and enhanced trust among users and content providers, as per the decentralization ethos articulated by \cite{benker2006wealth}.

\subsection{Societal and Ethical Implications}

The widespread adoption of systems like Viz could significantly impact labor markets, education, and information dissemination, necessitating a societal discourse on the role and regulation of AI.

While decentralization offers numerous benefits, it also raises ethical concerns, such as the potential for unchecked dissemination of biased or harmful AI models, necessitating robust governance structures.

\subsection{Challenges and Limitations}

Despite the efficiencies introduced by QLoRA, the computational demands of LLMs remain a challenge, particularly for smaller entities with limited resources.

Navigating the legal and ethical landscape, especially concerning copyright compliance and AI ethics, remains a complex and evolving challenge, as discussed by \cite{lessig2006code} in the context of internet regulation.

\subsection{Future Developments and Research Directions}
Continuous research in AI finetuning, like advancements beyond QLoRA, could further optimize model efficiency and effectiveness.

Exploring the potential of decentralizing aspects of the Viz system, such as using blockchain technology for transparent and secure transactions in the marketplace, could address issues like data privacy and user trust, as suggested by \cite{tapscott2016blockchain} in their exploration of blockchain revolution.

\section{Conclusion}

The purpose of this paper is to provide a thorough introduction and analysis of the Viz system. The Viz system is an innovative architecture that combines Quantized Low Rank Adapters (QLoRA) to fine-tune large language models (LLMs) in a way that is both legally compliant and economically feasible. This system makes a noteworthy contribution to the field of artificial intelligence by addressing the issues of computational efficiency, legal compliance, and economic sustainability when using LLMs. The following are the distinct contributions of this paper:

\begin{itemize} 
\item \textbf{Innovative Incorporation of QLoRA} The article emphasizes the unique utilization of QLoRA in Viz, illustrating how this method significantly reduces the computational resources needed for fine-tuning LLMs while still maintaining high performance levels. This contribution is particularly relevant in the context of making advanced AI technologies more accessible to a wider range of users and developers. 
\item \textbf{Creation of a Sustainable Economic Model} By proposing a marketplace for fine-tuned AI models, the article outlines an economic model that benefits content creators, AI developers, and end-users. This model not only fosters a competitive and innovative environment but also addresses the economic challenges traditionally associated with AI development and deployment.
\item \textbf{Legal and Ethical Framework} The article contributes to the discussion on legal and ethical considerations in AI by explaining how the Viz system adheres to copyright laws, data privacy regulations, and ethical AI principles. This aspect of the system is particularly important in an era where the societal and legal impacts of AI are increasingly scrutinized.
\item \textbf{Potential for Decentralization} The discussion on incorporating decentralization into the Viz system introduces a forward-thinking perspective on how blockchain technology and decentralized governance could enhance transparency, data security, and user trust in AI marketplaces.  
\end{itemize}

As the field of artificial intelligence (AI) progresses, systems such as Viz are positioned to have a significant impact on the future of AI development and application. Viz stands out with its unique methods for refining models, designing marketplaces, and ensuring legal compliance. This system sets a precedent for future advancements in the field, as it combines technological innovation, economic insight, and legal caution. By addressing current challenges and paving the way for future explorations in AI technology, Viz represents a convergence of various factors that shape the landscape of AI.

To sum up, the Viz system demonstrates a successful merging of technology, economy, and law, providing a holistic answer to the intricate issues present in the current AI environment. It serves as evidence of the possibilities that arise from collaborative and inventive methods in utilizing AI for the advantage of various stakeholders.

\bibliographystyle{unsrtnat}
\bibliography{references}

\end{document}